\documentclass{article}
\usepackage[nonatbib,final]{neurips_2019}
\usepackage[utf8]{inputenc} % allow utf-8 input
\usepackage[T1]{fontenc}    % use 8-bit T1 fonts
\usepackage{hyperref}       % hyperlinks
\usepackage{url}            % simple URL typesetting
\usepackage{booktabs}       % professional-quality tables
\usepackage{amsfonts}       % blackboard math symbols
\usepackage{nicefrac}       % compact symbols for 1/2, etc.
\usepackage{microtype}      % microtypography
\usepackage{color}
\usepackage{epsfig}
\usepackage{graphicx}
\usepackage{amsmath}
\usepackage{amssymb}
\usepackage{multirow}
\usepackage{verbatim}
\usepackage{caption}
\usepackage{array, siunitx, boldline, hhline}
\usepackage{threeparttable}
\usepackage{hyperref}
\usepackage[dvipsnames]{xcolor}
\definecolor{burgundy}{rgb}{0.5, 0.0, 0.13}
\definecolor{carnelian}{rgb}{0.7, 0.11, 0.11}

\title{Solution for the EPO CodeFest on Green Plastics: \\ Hierarchical multi-label classification of patents relating to green plastics using deep learning}

\author{
  Tingting Qiao$^{1}$\thanks{indicates equal contribution.} \quad Gonzalo Moro Pérez$^1$\footnotemark[1] \\
   \\
  $^1$Team GreenHands\\
    \\
  \small\texttt{tt.qiao.nl@gmail.com, gmoroperez@epo.org}\\}

\begin{document}
\maketitle

\begin{abstract}
This work aims at hierarchical multi-label patents classification for patents disclosing technologies related to green plastics. This is an emerging field for which there is currently no classification scheme, and hence, no labeled data is available, making this task particularly challenging. We first propose a classification scheme for this technology and a way to learn a machine learning model to classify patents into the proposed classification scheme. To achieve this, we come up with a strategy to automatically assign labels to patents in order to create a labeled training dataset that can be used to learn a classification model in a supervised learning setting. Using said training dataset, we come up with two classification models, a SciBERT Neural Network (SBNN) model and a SciBERT Hierarchical Neural Network (SBHNN) model. Both models use a BERT model as a feature extractor and on top of it, a neural network as a classifier. We carry out extensive experiments and report commonly evaluation metrics for this challenging classification problem. The experiment results verify the validity of our approach and show that our model sets a very strong benchmark for this problem. We also interpret our models by visualizing the word importance given by the trained model, which indicates the model is capable to extract high-level semantic information of input documents. Finally, we highlight how our solution fulfills the evaluation criteria for the EPO CodeFest and we also outline possible directions for future work. Our code has been made available at \url{https://github.com/epo/CF22-Green-Hands}.
\end{abstract}

\section{Introduction}

%Introduction to patents and patent classification
A patent is a type of intellectual property that gives its owner the legal right to exclude others from making, using, or selling an invention for a limited period of time in exchange for publishing an enabling disclosure of the invention. Soon after filing, patent applications are classified following a classification scheme. The International Patent Classification (IPC) and the Cooperative Patent Classification (CPC), which is a more specific version of the IPC, are two of the most commonly used classification schemes. Both the IPC and the CPC are hierarchical classification schemes and patents are classified as deeply as possible, i.e. there is no double classification in “parent” and “child” classes.

%Introduction to the motivation
It goes without saying that performing this classification manually by humans, e.g. by patent examiners, as it is currently done, is a very time-consuming task. Therefore, it would be very beneficial to be able to perform this classification automatically and there is an increasing amount of effort in using machine learning models for this purpose. 

%Additional challenges
An additional challenge is that the classification schemes are not static. As technology evolves, some technical areas become obsolete while new ones appear. To reflect this, the IPC and CPC are revised periodically and, if needed, updated. Upon updating the classification scheme, classification models also need to be updated to take into account the changes. Just to provide an example, in 2022, several “child” classes were added to the class G06N10/00 for quantum computing, namely G06N10/20, G06N10/40, G06N10/60, G06N10/70 and G06N10/80. This adds a layer of complexity on top of the patent classification task as there is no training data available for the new classes. The obvious solution is to manually check all the patents classified in G06N10/00, assign “child” classes as necessary, and subsequently, update the classification model using this manually labeled data. However, this is a very time-consuming approach and it would be very beneficial to find an alternative way to update the classification model.

%Green plastics
In this work, we focus on technologies relating to green plastics. The term "green plastics" refers to a way of achieving a more circular plastics industry, for example by plastics with a reduced or minimized environmental impact or by processes for improved plastics recycling and minimizing plastic waste. In the last couple of decades, the amount of patents disclosing technologies related to green plastics has increased considerably. In order to make the knowledge contained in said patents more readily available to everybody, there is a need for having a patent classification scheme relating to this technology.

%What do we do? Contributions?
We propose a classification scheme, based on \cite{EPOgreenplastics}, for technologies relating to green plastics and an automatic way to assign weak labels to patents so that a machine learning model, based on BERT as a feature extractor and a neural network as a classifier, can be learned in a supervised setting. To the best of our knowledge, this is the first work that applies a weak supervision strategy for the classification of patents in a new classification scheme.

%Final paragraph
The paper is organized as follows: First, related work is briefly discussed. Second, the classification scheme is introduced and a labeled training dataset is obtained. Consequently, a machine learning model is proposed and trained on the training dataset. Subsequently, several experiments are carried out and the results are reported and discussed. Finally, the fulfillment of the EPO CodeFest criteria is presented and future work is thoroughly discussed.

\section{Related work}

% Patent classification using traditional machine learning models
In the last decades, there is a large number of works focusing on building machine learning models for patent classification. Earlier works used traditional machine learning methods such as k-Nearest neighbors (k-NN) \cite{fall2003automated}, support vector machine (SVM) \cite{fall2003automated, wu2010patent, d2013text}, Naive Bayes (NB) \cite{fall2003automated, d2013text}, k-means clustering \cite{kim2008visualization} and artificial neural networks \cite{trappey2006development, guyot2010myclass}. 

% Patent classification using deep learning
In the past decade, deep learning techniques have also been applied to patent classification outperforming previous methods. \cite{li2018deeppatent} proposed a deep learning algorithm based on pre-trained word embeddings and convolutional neural networks (CNN). The input to the models consists of the title and the abstract of the patent and the output consists of one or multiple classes at the sub-class label. \cite{huang2019hierarchical} proposed using BiLSTM initialized with word2vec and a hierarchical attention-based memory unit. The input to the model consists of the title and the abstract of the patent and the output consists of a probability value for each class in the hierarchy.

% Patent classification with transformers
Recently, transformer-based neural language models such as BERT \cite{devlin2018bert} have outperformed previous state-of-the-art models in several natural language processing tasks. Unsurprisingly, BERT has also been applied to patent classification. \cite{lee2020patent} proposed to fine-tune a pre-trained BERT model for patent classification. Very recently, \cite{pujari2021multi} proposed to fine-tune a pre-trained BERT model combined with a neural network based hierarchical classifier. In particular, they used SciBERT \cite{beltagy-etal-2019-scibert}, which is a BERT model trained on a corpus of scientific publications and is hence closer to the patent domain.

% What is the output?
It is worthwhile to point out that most of these works focus on learning a model for classifying patent documents locally at a single hierarchical level, i.e. only “child” classes are assigned by the model, as e.g. \cite{li2018deeppatent} or \cite{lee2020patent}, or globally by treating all labels independently, e.g. \cite{huang2019hierarchical}. However, recent work \cite{pujari2021multi} proposes a hierarchical neural network classifier with as many output heads as classes in the hierarchy. 

% Prior-art works assume the existence of labels
All prior-art works mentioned above assume the existence of a labeled training dataset, i.e. they perform patent classification based on an already existing classification scheme. This is however not the case in our work as we first need to propose a classification scheme and subsequently, we learn a classification model for the proposed classification scheme.

% Paragraph about weak supervision
Weak supervision is an approach for machine learning in which noisy or imprecise labels are provided to learn a model in a supervised learning setting. There are plenty of works using weak supervision in different natural language processing tasks, such as text classification. Recently, \cite{weak_super} proposed multiple weak supervision strategies to label text data automatically. One of the strategies proposed is based on heuristic rules consisting of regular expression matching.

\section{Classification scheme}

\begin{figure}[htb!]
\centering
\noindent\includegraphics[width=1\columnwidth]{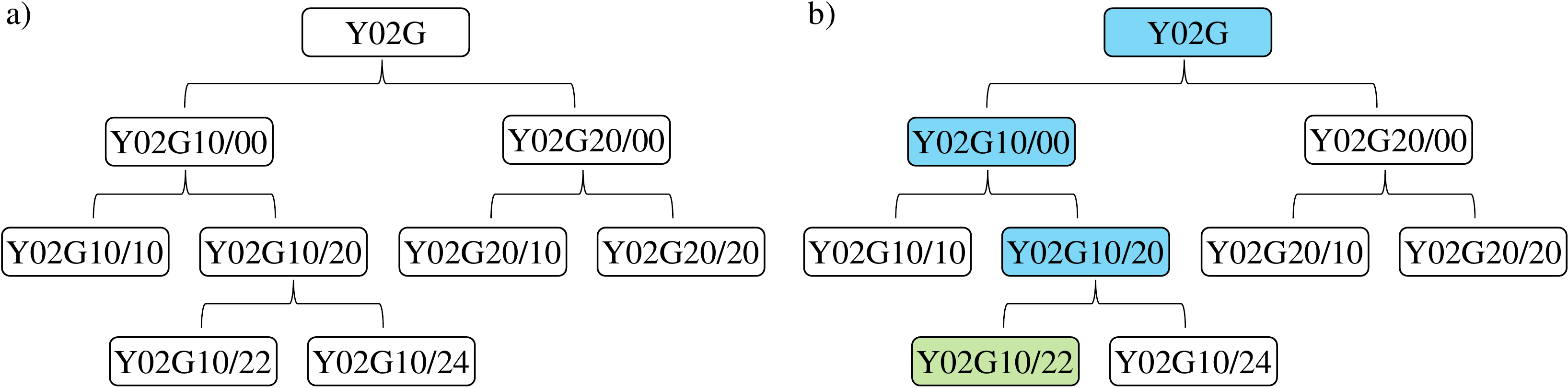}
\captionsetup{font={footnotesize}}
\protect\caption{\label{fig:class_hierarchy} a) Proposed classification scheme for technologies related to green plastics. b) Hierarchical labels: A patent belonging to class Y02G10/22 (shown in green) is considered to also belong to all the ancestor classes (shown in blue) and it is given the label \textit{l} = [\textcolor{Cyan}{1}, \textcolor{Cyan}{1}, 0, \textcolor{Cyan}{1}, \textcolor{LimeGreen}{1}, 0, 0, 0, 0]; as opposed to merely \textit{l} = [0, 0, 0, 0, \textcolor{LimeGreen}{1}, 0, 0, 0, 0]. Note that the label list corresponds to the class list [Y02G, Y02G10/00, Y02G10/10, Y02G10/20, Y02G10/22, Y02G10/24, Y02G20/00, Y02G20/10, Y02G20/20]. Best viewed in color.}
%\vspace{-0.45cm}
\end{figure}

As discussed above, there is no existing classification scheme for technologies relating to green plastics. Therefore, a classification scheme, based on \cite{EPOgreenplastics}, is proposed as shown in Figure \ref{fig:class_hierarchy}a. The class definitions are summarized in Table \ref{table:classification}. The root node is Y02G for technologies relating to green plastics, which is divided into Y02G10/00 for recycling plastic waste and Y02G20/00 for alternative plastics. Y02G10/00 is in turn divided into Y02G10/10 for plastic waste recovery, including collecting, sorting, separating and cleaning plastic waste; and Y02G10/20 for plastic waste recycling, including recycling methods such as plastic-to-compost, plastic-to-monomer, plastic-to-incineration and plastic-to-energy. Y02G10/20 also has two child classes, which are Y02G10/22 for plastic-to-product recycling, including mechanical recycling, such as melting and reforming thermoplastics, and Y02G10/24 for plastic-to-feedstock recycling, such as cracking, gasification and pyrolysis. Y02G20/00 is divided into Y02G20/10 for bioplastics and Y02G20/20 for designs for easier recycling.

\section{Solution design}

Now that we have a classification scheme, we need to propose a way to classify patents. An option could be to assign keywords to classes and, if any of those keywords are present in a patent, assign the corresponding classes to the patent accordingly. However, such a simple approach has several weaknesses. It is limited to the specified keywords and cannot extend to any synonyms which have not been explicitly specified. Moreover, it is unable to find semantically equivalent expressions which do not match the specified keywords. 

However, machine learning models have the capacity of learning high-level semantic representations from data without the need for domain experts. For this reason, we decided to propose a machine learning model. 

However, at this point, we do not have a labeled training dataset for our classification scheme. Learning a machine learning classification model in an unsupervised setting (when the training dataset is unlabeled) is not feasible as it would not be possible to control the classes that the model learns. Therefore, we decided to learn a classification model in a supervised setting (when the training dataset is labeled). In the following sections, we explain how we build our labeled training dataset and how we obtain our classification models. 

\begin{table}[htb!]
 \renewcommand{\arraystretch}{1.4}
 \begin{tabular}{|l|l|p{78mm}|}
        \hline
       \textbf{Class} & \textbf{Definition} & \textbf{Keywords}\\
        \hline
        Y02G  & Green plastics & green+ 4d plastic+\\
        \hline
        Y02G10/00 & Recycling of plastic waste & recycl+ 4d plastic+\\
        \hline
        Y02G10/10 & Plastic waste recovery & (plastic+ 3d wast+) or ((collect+ or sort+ or separat+ or clean+) 6d plastic+) \\
        \hline
        Y02G10/20 & Plastic waste recycling & ((recycle+ 4d plastic+) 20d (compost+ or fertili+)) or ((recycle+ 4d plastic+) 20d (depolymer+ or repolymer+)) or ((recycle+ 4d plastic+) 4d incinerat+) \\
        \hline
        Y02G10/22 & Plastic-to-product & (plastic+ 4d recycle+) 20d (+melt+ or extrud+ or pellet+) \\
        \hline
        Y02G10/24 & Plastic-to-feedstock & ((feedstock+ 2d recycl+) 20d plastic+) \\
        \hline
        Y02G20/00 & Alternative plastics & alternativ+ 2d plastic+ \\
        \hline
        Y02G20/10 & Bioplastics & bioplastic+ or ((biolog+ or biodegrad+ or biobased+ \newline or compostable+) 4d plastic+) \\
        \hline
        Y02G20/20 & Designs for easier recycling & vitrimer+ or ((covalent+ 2d adapt+) 2d net+) or \newline ((selfheal+ or selfrepair+) 2d polymer+) 20d recycl+ \\
        \hline
        \end{tabular}
\caption{Definition and keywords for each class in the proposed classification scheme. The symbol "+" means none or more characters. The expression "$w_1$ \textit{n}d $w_2$" means that between words $w_1$ and $w_2$ there may be $n$ other words and the relative order in the sequence of words between $w_1$ and $w_2$ is irrelevant.}
\label{table:classification}
\end{table}

\section{Training dataset}

To learn a machine learning classification model in a supervised setting, we need to prepare a labeled training dataset \textit{X} = \{(\textit{$P_1$}, \textit{$l_1$}), (\textit{$P_2$}, \textit{$l_2$}), ..., (\textit{$P_M$}, \textit{$L_M$)}\}, where each patent \textit{$P_i$} consists of a sequence of \textit{N} words, i.e. \textit{$P_i$} = \{\textit{$w_1$}, \textit{$w_2$}, ..., \textit{$w_N$}\}, and is associated with a label \textit{l}. Ideally, the labels would be assigned to the patents manually by humans. However, this is a very time-consuming task and therefore, we explore a way to provide weak labels in an automatic manner.  

\subsection{Raw dataset}

To build the training dataset, we use the \texttt{EP full text data} \cite{office} which contains all publications of EP patent applications and patent specifications from 1978 until the end of January 2022. The dataset comprises over 6 million publications, each comprising several fields such as title, abstract, description, claims, language, filing date, publication date, classes, search report, etc. The entire dataset is about 260 gigabytes in size. 

EP patents may be published in English, French or German. Since available pre-trained language models are usually monolingual \cite{devlin2018bert, beltagy-etal-2019-scibert}, we select only patents written in English. For each patent, the title, the abstract and the description are obtained while the other fields are discarded. Some standard text pre-processing steps, such as removing punctuation marks and stop-words, minimizing capital letters and tokenization, are carried out. 

This dataset is unlabeled since the patents are not assigned classes in the classification hierarchy described in Figure \ref{fig:class_hierarchy}a. In the following, we explain the structure of the labels and how to assign them to build a labeled training dataset.

\subsection{Hierarchical label definition}
\label{subsec:hierarchical_labels}

According to our classification scheme, there are nine classes and hence, each patent will be assigned a label \textit{l} consisting of a 9-dimensional binary vector, where 1 means that the patent belongs to the corresponding class and 0 means that the patent does not belong to the corresponding class. The labels look as follows: \textit{l} = [Y02G, Y02G10/00, Y02G10/10, Y02G10/20, Y02G10/22, Y02G10/24, Y02G20/00, Y02G20/10, Y02G20/20], where each element is a binary value associated with the corresponding class. 

In order to provide more informative labels to the model for training, we propose to use hierarchical labels, meaning that for a patent belonging to a certain class, we also assign all the ancestor classes of said class. The idea behind this is that the model should learn that the given patent belongs to all the corresponding ancestors as well. This is illustrated in Figure \ref{fig:class_hierarchy}b. 

\subsection{Labeling process}

\begin{figure}[h!]
\centering
\noindent\includegraphics[width=0.6\columnwidth]{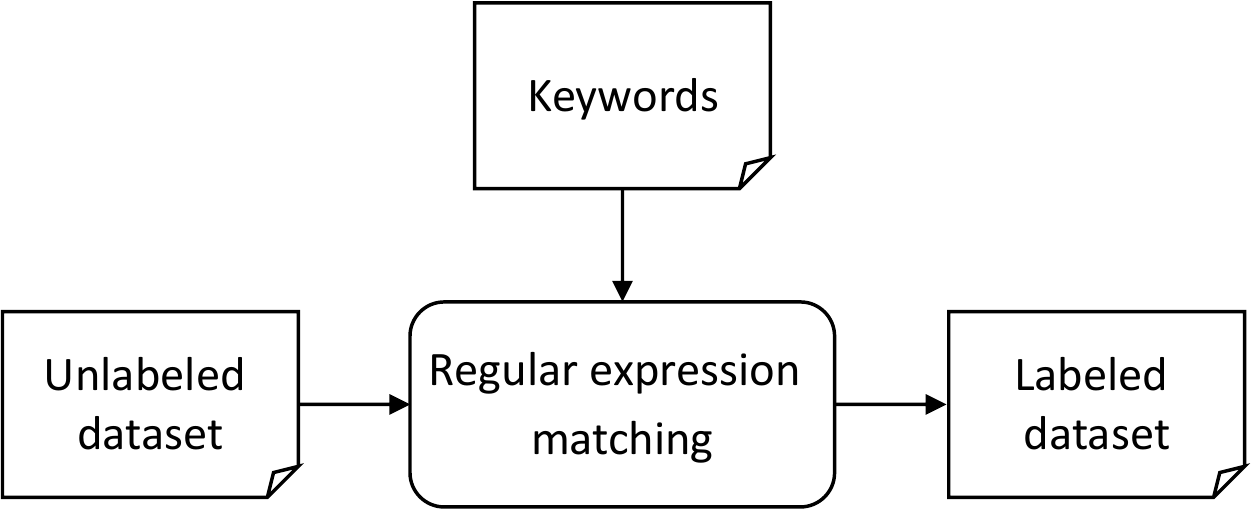}
\captionsetup{font={footnotesize}}
\protect\caption{\label{fig:weak_super}Schematic representation of the labeling process.}
%\vspace{-0.45cm}
\end{figure}

To assign the labels to the patents in an automatic way, we first define the keywords to each class. The list of keywords is provided in Table \ref{table:classification}. Consequently, we translate the keywords into regular expressions and search for them in the description of each patent. If the keyword is found a \textit{k} number of times, the patent is assigned to the class corresponding to the found keyword. This process is represented schematically in Figure \ref{fig:weak_super}.

\subsection{Building the training dataset}

\begin{figure}[h!]
\centering
\noindent\includegraphics[width=1\columnwidth]{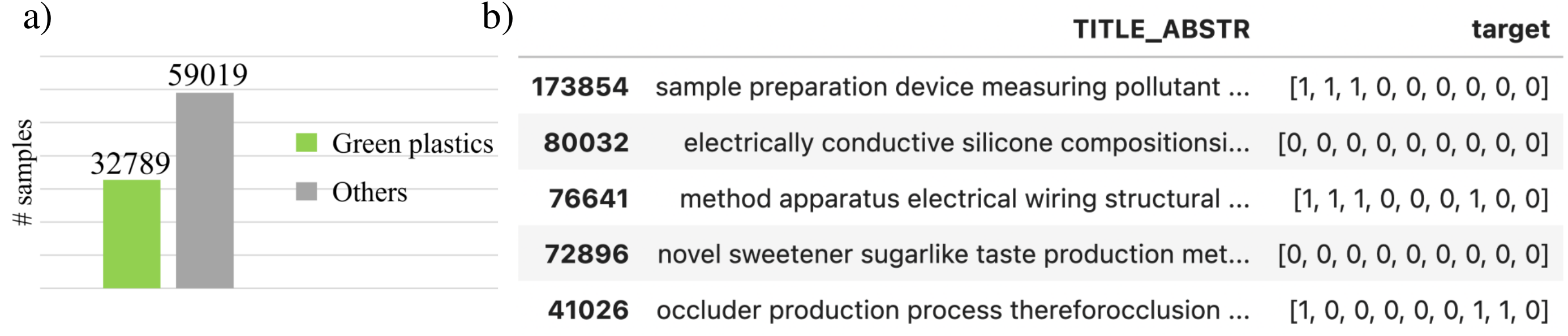}
\captionsetup{font={footnotesize}}
\protect\caption{\label{fig:sample_dataset}a) Number of patents belonging to green plastics and number of patents belonging to other fields of technology. b) Sample from the labeled training dataset. After labeling, the description is dropped and only the title and the abstract are used as input for the classification model.}
%\vspace{-0.45cm}
\end{figure}

The labeling process described in Figure \ref{fig:weak_super} is then performed to all the raw data that survived our raw data quality check and filters, e.g. English files, without missing content etc. In order to obtain a labeled training dataset. we sample all patents which have been labeled as belonging to green plastics (positive samples) and around two times more patents which have not been labeled as belonging to green plastics (negative samples). For the negative samples, we randomly sample patents related to any other field of technology. Specifically, in a similar way as for the patents related to green plastics, we also sample patents related to conventional plastics in our negative sample set, so that the classification model can learn to differentiate between conventional plastics and green plastics. The number of patents belonging to green plastics and the number of patents belonging to other fields of technology are shown in Figure \ref{fig:sample_dataset}a. It can be see that the constructed dataset is imbalanced meaning that the samples of one class (in this case, green plastics) are considerably lower than the samples of other classes. This is done on purpose to mimic the real situation and also to challeng model to learn better parameters. A sample of the labeled training dataset can be seen in Figure \ref{fig:sample_dataset}b, where the column TITLE\_ABSTR is the input which is a combination of the title and the abstract of one patent file, the target column is to indicate if the patent belongs to certain classes, 1 means yes. 

We split the resulting labeled dataset into a training set, a validation set and a test set. The breakdown of the number of patents for each class in each of these sets is shown in Table \ref{table:dataset}. This table shows the number of positive samples (+) and negative samples (-) for each class. As can be seen, the dataset is heavily imbalanced, above all for some classes that are deeper in the class hierarchy, such as Y02G10/24 or Y02G20/20, which have less than 100 positive samples.

At this point, it could be possible to try to obtain more positive samples for those classes by enlarging the set of keywords in Table \ref{table:classification}. However, expert knowledge would be required to suggest more relevant keywords. Another possibility could be to ignore classes that have less than a threshold number of samples (e.g. 100 or 500), since it will be very hard for the machine learning model to learn those classes given the large imbalance. However, we decided to continue with the dataset of Table \ref{table:dataset}. In the evaluation section, we discuss how we believe that this imbalance could be tackled. Moreover, we also report results for each level of the classification hierarchy level and for each class, which should not be considerably affected by the presence of these heavily imbalanced classes during training.

\begin{table}[htb!]
\centering
 \renewcommand{\arraystretch}{1.4}
 \begin{tabular}{|l|r r|r r|r r|}
        \hline
        \multicolumn{1}{|c|}{\textbf{Class}}  &
        \multicolumn{2}{c|}{\textbf{Training set}}  &
        \multicolumn{2}{c|}{\textbf{Validation set}}&
        \multicolumn{2}{c|}{\textbf{Test set}} \\
         & + & - & + & - & + & - \\ 
        \hline
        Y02G  & 26286 & 47160 & 3244 & 5937 & 3259 & 5922  \\
        \hline
        Y02G10/00 & 19106 & 54340 & 2335 & 6846 & 2376 & 6805 \\
        \hline
        Y02G10/10 & 17452 & 55994 & 2150 & 7031 & 2171 & 7010 \\
        \hline
        Y02G10/20 & 494 & 72952 & 63 & 9118 & 61 & 9120 \\
        \hline
        Y02G10/22 & 400 & 73046 & 50 & 9131 & 51 & 9130 \\
        \hline
        Y02G10/24 & 33 & 73413 & 3 & 9178 & 4 & 9177 \\
        \hline
        Y02G20/00 & 7755 & 65691 & 977 & 8204 & 946 & 8235 \\
        \hline
        Y02G20/10 & 2901 & 70545 & 375 & 8806 & 376 & 8805 \\
        \hline
        Y02G20/20 & 10 & 73436 & 0 & 9181 & 1 & 9180 \\
        \hline
        \hline
        \multicolumn{1}{|c|}{Total}  &
        \multicolumn{2}{c|}{73446}  &
        \multicolumn{2}{c|}{9181}&
        \multicolumn{2}{c|}{9181} \\
        \hline
        \end{tabular}
\caption{Number of patents for each class in the training, validation and test sets. This is obtained with a threshold \textit{k} = 1. "+" indicates positive samples, i.e. patents belonging to the corresponding class, and "-" indicates negative samples, i.e. patents not belonging to the corresponding class. Patents counted for a given class are also counted for all the ancestors of that class. For example, if a patent is counted for Y02G10/22, it is also counted for Y02G10/20, Y02G10/00 and Y02G, in line with the labeling process described in Figure \ref{fig:class_hierarchy}b. The last row indicates the total amount of patents in each of the training, validation and test sets.}
\label{table:dataset}
\end{table} 

\section{Classification model}

Our goal is to learn a classification model $\Omega$ with parameters $\Theta$ such that given a patent $P$, it returns a 9-dimensional classification result $y$, where each element in $y$ corresponds to the probability of the patent $P$ belonging to the corresponding class in the classification scheme of Figure \ref{fig:class_hierarchy}a. Mathematically,

\begin{equation*}
\small
\textit{y} = \Omega(\textit{P}; \Theta); 
\end{equation*}

\subsection{Architecture}

\begin{figure}[htb!]
%\hspace*{-1in}
\centering
\noindent\includegraphics[width=1\columnwidth]{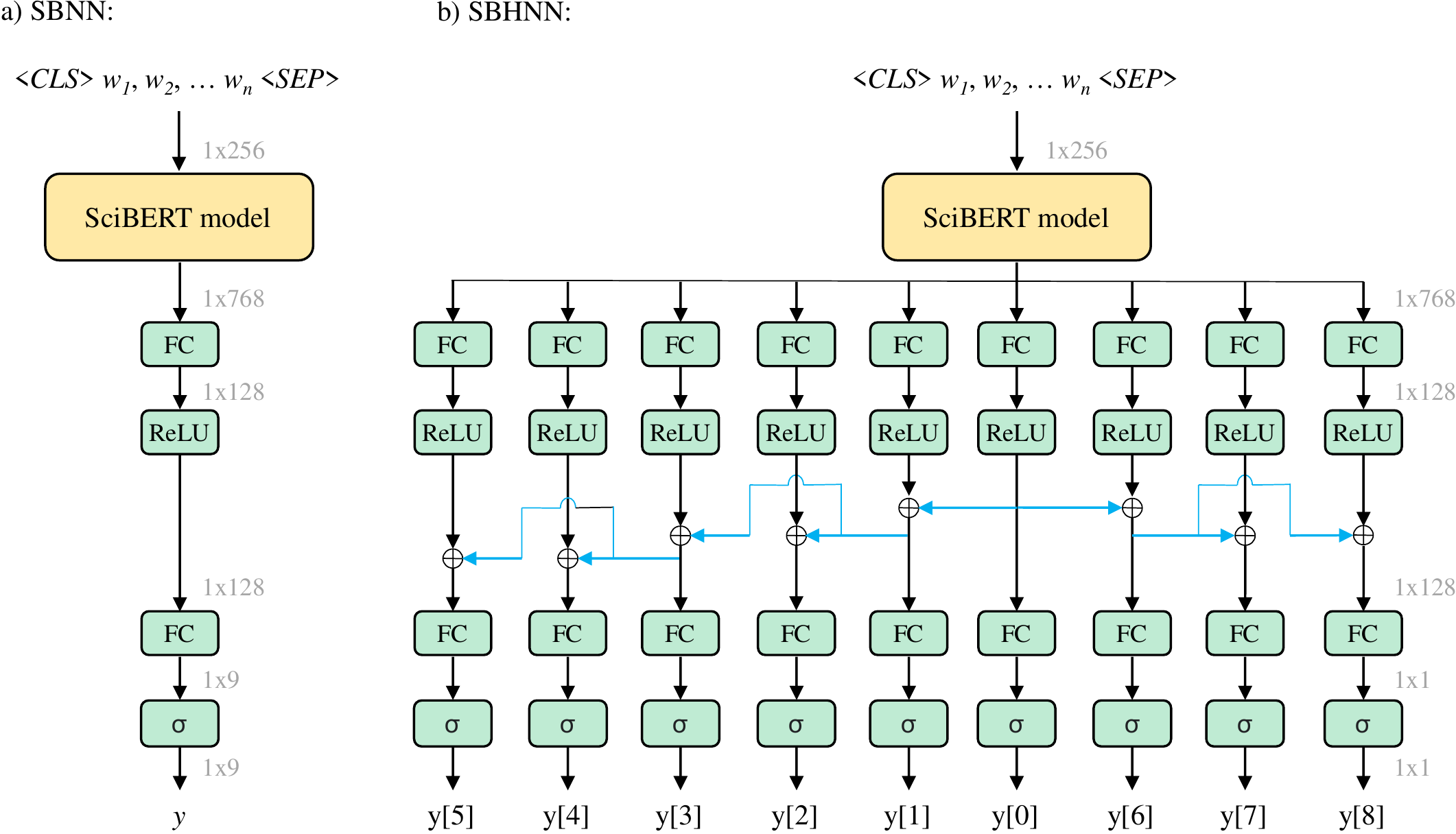}
\captionsetup{font={footnotesize}}
\protect\caption{\label{fig:model} a) Architecture of SBNN. b) Architecture of SBHNN. The feature extractor is displayed in orange and the classifier is displayed in green. The blue arrows indicate the hierarchical connections between classification heads. The sizes of the vectors are shown in grey. Best viewed in color.}
%\vspace{-0.45cm}
\end{figure}

We propose two classification models as shown in Figure \ref{fig:model}. Both models use a BERT model as a feature extractor and a neural network as a classifier. In the first model, SBNN, the classifier is a conventional neural network and in the second model, SBHNN, the classifier is a hierarchical neural network. Similar to \cite{pujari2021multi}, the BERT model is a pre-trained SciBERT model \cite{beltagy-etal-2019-scibert}. The input to the model is a 256-dimensional vector consisting of a concatenation of the patent’s title and abstract, as done in \cite{li2018deeppatent, huang2019hierarchical, lee2020patent, pujari2021multi} and the output is a 768-dimensional feature vector $h$ corresponding to the CLS embedding, as done in \cite{lee2020patent, pujari2021multi}:

\begin{equation*}
\small
h = SciBERT(\textit{$w_1$}, \textit{$w_2$}, ... , \textit{$w_N$}; \Theta_B); 
\end{equation*}

where $\Theta_B$ represents the parameters of SciBERT.

The feature vector $h$ is subsequently input to a neural network classifier $NN$ which outputs a classification result: 

\begin{equation*}
\small
y = NN(h; \Theta_N); 
\end{equation*}

where $\Theta_N$ represents the parameters of the neural network. 

The neural network is implemented by cascading fully connected layers (FC), which implement a linear matrix multiplication and add a bias term, and a non-linear activation function, e.g. ReLU or sigmoid, $\sigma$. Mathematically, a fully connected layer with a non-linear activation function is implemented as follows: 

\begin{equation*}
\small
t = a(Wx + b); 
\end{equation*}

where $x$ and $t$ are the input and output, $W$ is a learnable weight matrix, $b$ is a learnable bias vector and $a(z)$ denotes a non-linear function, such as:

\begin{equation*}
\small
ReLU(z) = max\{0, z\};  \quad \quad \sigma (z) = \frac{1}{1 + e^{-z}}
\end{equation*}

In SBNN, the neural network classifier consists of a fully connected layer with ReLU activation followed by another fully connected layer with sigmoid activation. In SBHNN, inspired by \cite{pujari2021multi}, the neural network classifier consists of a hierarchical neural network with one classification head per class in which the connections between different classification heads, implemented as element-wise additions, directly reflect the classification scheme of Figure \ref{fig:class_hierarchy}a. 

\subsection{Loss function}

The model $\Omega$ is trained using the labeled training dataset \textit{X} to minimize the binary cross-entropy loss, which, for a single patent, is expressed as follows: 

\begin{equation*}
\small
L = - \sum_{i=0}^{C=8} \beta_i [\gamma_i \cdot l_i \cdot log(y_i) + (1 - l_i) \cdot log(1 - y_i)]; 
\end{equation*}

where $C$ represents the number of classes, $\beta$ is a class importance weight (if $\beta$ = [1, 1, 1, 1, 1, 1, 1, 1, 1], all classes are given the same importance) and $\gamma$ is a positive sample weight to compensate the imbalanced training set, and in the end, it will be reflected as a trade-off between precision and recall (if, for a given class $c$, $\gamma_c$ = 1, false positives and false negative are given the same importance). 

During training, we update the parameters $\Theta_B$ and $\Theta_N$ to minimize this loss function by mini-batch gradient descent with Adam optimizer \cite{kingma2014adam}.

\section{Evaluation}

We evaluate the performance of our models on our training dataset based on several evaluation metrics which are commonly used in hierarchical text classification tasks. In the following, we report and analyze the results.

\subsection{Evaluation metrics}

Prior-art \cite{pujari2021multi} used \textit{hierarchical} precision, \textit{hierarchical} recall and \textit{hierarchical} F1-score as proposed by \cite{kiritchenko2005functional}. These metrics are more suitable for hierarchical classification tasks than the conventional precision, recall and F1 score as they give credits to partially correct classifications and discriminate errors by both distance and depth in the classification hierarchy. These are defined as follows:

\begin{equation*}
\small
hP = \frac{\sum_{i}^{} |Y_i \cap L_i| }{\sum_{i}^{}|L_i|}; \quad \quad hR = \frac{\sum_{i}^{}|Y_i \cap L_i|}{\sum_{i}^{}|L_i|}; \quad \quad hF1 = 2 \cdot \frac{hP \cdot hR}{hP + hR}; 
\end{equation*}

wherein, for each test instance $i$, the set $Y_i$ consists of all predicted labels and their respective ancestors, the set $L_i$ consists of all true labels including ancestors, $|\cdot|$ denotes the cardinality of a set and $\cap$ denotes the intersection of sets. Hence, if for a given patent the predicted class is Y02G10/20 while the target class is Y02G10/22, the set $Y_i$ = \{Y02G10/20, Y02G10/00, Y02G\} and the set $L_i$ = \{Y02G10/22, Y02G10/20, Y02G10/00, Y02G\}. The patent is correctly assigned the classes from the extended set $|Y_i \cap L_i|$ = 3, there are $|Y_i|$ = 3 assigned classes and $|L_i|$ = 4 target classes. Therefore, we get $hP = \frac{3}{3}$ and $hR = \frac{3}{4}$.

As done by \cite{pujari2021multi}, we report per instance $macro$-scores, that compute the scores independently per class and then average them, as well as $micro$-scores, that aggregate the contributions of each class to compute an global average score. Both are implemented with scikit-learn \cite{scikit-learn}. Unless otherwise stated, the decision threshold is set to 0.5, meaning that we consider that a patent is classified to class if the probability of the classification model output, i.e. after the sigmoid layer, is at least 0.5. 

In addition, we compute the Area Under the Precision-Recall Curve (AUPRC) \cite{davis2006relationship}, which does not require defining a decision threshold. We also compute the accuracy, even though this is not the best metric when the dataset is imbalanced. We use the implementation from scikit-learn \cite{scikit-learn} to compute both the AUPRC and the accuracy.

\subsection{Implementation details}

We build the labeled training dataset using the \texttt{EP full text data} \cite{office} as indicated in Figure \ref{fig:weak_super}. We download and store all the data in a Toshiba Canvio Basics USB 3.0 Hard Drive with 1 TB of storage. Consequently, we label and sample the training dataset in a MacBook Pro with a 2,7 GHz Dual-Core Intel Core i5.

We implement our models in Python using Pytorch \cite{NEURIPS2019_9015}. We use the HuggingFace Transformers library \cite{huggingface} for instantiating SciBERT. We use the same hyper-parameters for training both classification models. We have a learning rate of 2e-6, a batch size of 96 and a dropout of 0.5. The class importance is set to $\beta$ = [4, 3, 2, 2, 1, 1, 3, 2, 2], and hence, we give more importance to classes that are higher up in the classification hierarchy. The positive sample weight is set to $\gamma$ = [2, 2, 2, 2, 2, 2, 2, 2, 2] to compensate for the class imbalance. The parameters of SciBERT are initialized to the pre-trained values and the parameters of the neural network are initialized randomly. 

\begin{figure}[htb!]
%\hspace*{-1in}
\centering
\noindent\includegraphics[width=1\columnwidth]{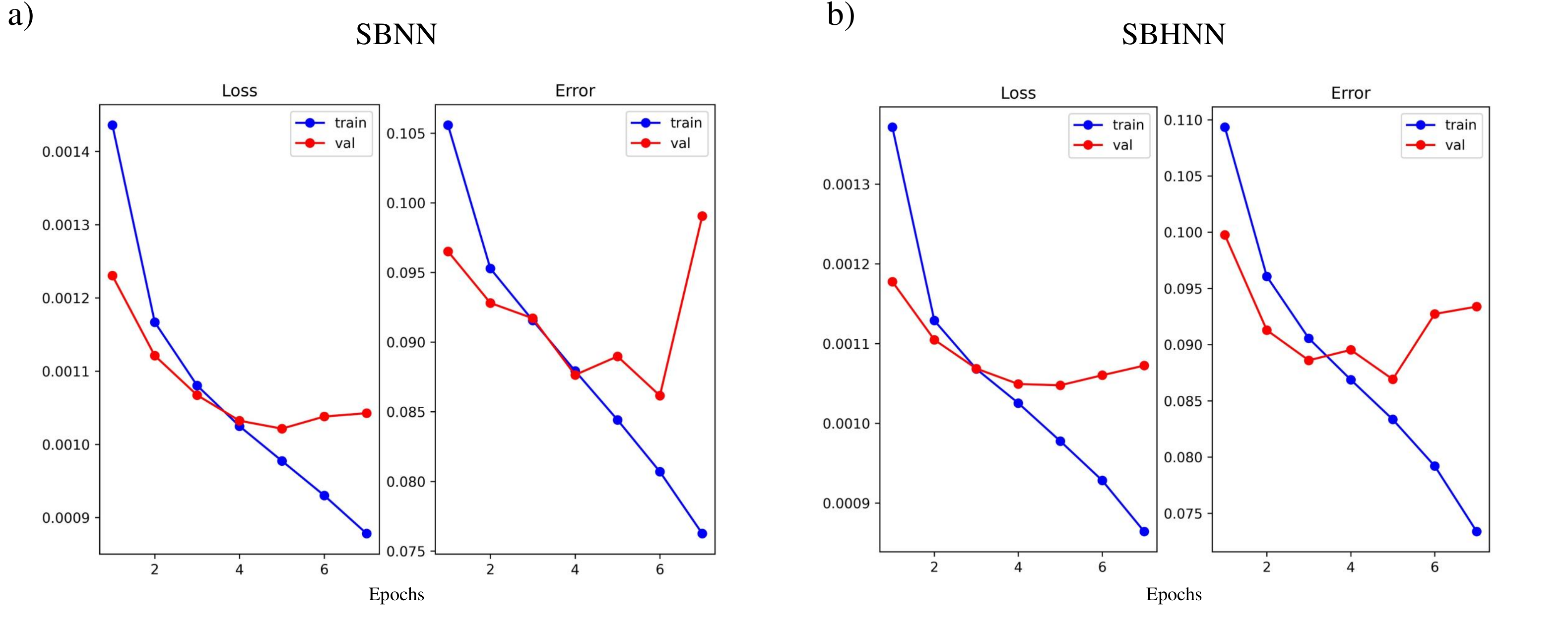}
\captionsetup{font={footnotesize}}
\protect\caption{\label{fig:loss_error} Evolution of the train and validation loss and error. Best viewed in color.}
%\vspace{-0.45cm}
\end{figure}

We train the models until the loss in the validation set stops decrease and the error stops to decrease in order to avoid overfitting. In Figure \ref{fig:loss_error}, we show the evolution of the train and validation loss and error. For SBNN, we chose the model after 6 training epochs; and for SBHNN, we chose the model after 5 epochs. The training of a single model takes approximately 1.2 hours on an Nvidia GeForce RTX 3090 with 24GB of GDDR6X memory.

\subsection{Experimental results}
\label{subsection:experimental_resuls}

We perform experiments to verify the validity of our classification models. Since we are dealing with a hierarchical multi-label classification problem with imbalanced data, in order to give a complete overview, we report results based on the whole classification hierarchy, based on each level in the classification hierarchy and based on each class.

\textbf{Performance for the whole classification hierarchy.} Table \ref{table:results_whole} shows the overall results obtained for SBNN and SBHNN. From Table \ref{table:results_whole} we can also see that SBHNN achieves better performance than SBNN in terms of both macro and micro F1 scores. For both models, the micro scores are considerably higher than the macro scores. This is due to the fact that macro scores calculate metrics for each class, and compute their unweighted mean, without taking class imbalance into account. On the other hand, micro scores calculate the metrics globally by considering each output independently. Therefore, they are relatively less biased towards the classes which have less positive training samples, and hence, provide higher scores. 

It is noticeable that the overall results of both models don't seem to be very satisfactory. This is actually quite commonly shown in the literature \cite{pujari2021multi}. The overall low performance is due to the challenging nature of the hierarchical classification problem. Moreover, in our case, the lower the class level is, the less number of positive samples we have, and therefore, the more challenging it is to train a good performing classification model. It should be kept in mind that the calculations of these evaluation metrics are negatively biased by these classes with insufficient data. 

\textbf{Performance for each level in the classification hierarchy.} Table \ref{table:results_per_level} shows the results per level obtained for SBNN and SBHNN. It is clearly shown that both SBNN and SBHNN have a better performance, both in precision and recall, the higher the class is in the classification hierarchy. For the first level (Y02G), the precision and recall of both models are higher than 70$\%$. For the second level (Y02G10/00 and Y02G20/00), they still reach 60-50$\%$ and for the third level (Y02G10/10, Y02G10/20, Y02G20/10 and Y02G20/20) and the fourth level (Y02G10/22 and Y02G10/24), they are around 40-30$\%$. This is to be expected as the number of positive training samples in lower levels is considerably lower, see Table \ref{table:dataset}, therefore, there is not much information for both models to learn from. Moreover, we also gave a higher class importance, i.e. $\beta$, to classes in higher levels in the classification scheme during the training process, so the model was trained to focus more on the classes which have more positive samples.  

We can also see that SBNN actually performs slightly better in level 1 in terms of both macro and micro F1 scores, whereas SBHNN obtains better results in lower levels. This is probably because of the unique architecture of SBHNN, in which there are independent classification heads for each class. The model can thus take advantage of the more complex neural network and learn better from the data.

\begin{table}[htb!]
\centering
 \renewcommand{\arraystretch}{1.4}
 \begin{tabular}{|l|c c c|c c c|c|c|}
        \hline
        \multicolumn{1}{|c|}{}  &
        \multicolumn{3}{c|}{\textbf{macro-avg.}}  &
        \multicolumn{3}{c|}{\textbf{micro-avg.}}&
        \multicolumn{1}{c|}{\textbf{AUPRC}} & 
        \multicolumn{1}{c|}{\textbf{Accuracy}} \\
        Model & hP & hR & hF1 & hP & hR & hF1 & &\\
        \hline
        SBNN  & 0.3587 & 0.2290 & 0.2584 & \textbf{0.6113} & 0.5087 & 0.5553 & \textbf{0.6317} & 0.6026\\
        \hline
        SBHNN & \textbf{0.4856} & \textbf{0.2807} & \textbf{0.3197} & 0.5882 & \textbf{0.5398} & \textbf{0.5629} & 0.6032 & \textbf{0.6087}\\
        \hline
        \end{tabular}
\caption{Evaluation metrics for the whole classification hierarchy.}
\label{table:results_whole}
\end{table}

\begin{table}[htb!]
\centering
 \renewcommand{\arraystretch}{1.3}
 \begin{tabular}{|l|c|c c c|c c c|c|c|}
        \hline
        \multicolumn{1}{|c}{}  &
        \multicolumn{1}{|c|}{} &
        \multicolumn{3}{c|}{\textbf{macro-avg.}}  &
        \multicolumn{3}{c|}{\textbf{micro-avg.}} &
        \multicolumn{1}{c|}{\textbf{AUPRC}} &
        \multicolumn{1}{c|}{\textbf{Accuracy}} \\
        Model & Level & hP & hR & hF1 & hP & hR & hF1 & & \\
        \hline
        \multirow{ 4}{*}{SBNN} & 1 & \textbf{0.7173} & 0.7160 & \textbf{0.7166} & \textbf{0.7412} & \textbf{0.7412} & \textbf{0.7412} & \textbf{0.7076} & \textbf{0.7412} \\
        & 2 & \textbf{0.6276} & 0.4436 & 0.4805 & \textbf{0.6198} & 0.5341 & 0.5738 & 0.6325 & \textbf{0.6580} \\
        & 3 & 0.4612 & 0.2944 & 0.3322 & \textbf{0.6113} & 0.5118 & 0.5571 & 0.6058 & \textbf{0.6317} \\
        & 4 & 0.3587 & 0.2290 & 0.2584 & \textbf{0.6113} & 0.5087 & 0.5553 & 0.6026 & \textbf{0.6317}\\
        \hline
        \multirow{ 4}{*}{SBHNN} & 1 & 0.7096 & \textbf{0.7208} & 0.7131 & 0.7294 & 0.7294 & 0.7294 & 0.7057 & 0.7294 \\
        & 2 & 0.5546 & \textbf{0.4873} & \textbf{0.5052} & 0.5890 & \textbf{0.5716} & \textbf{0.5802} & \textbf{0.6342} & 0.6269\\
        & 3 & \textbf{0.5172} & \textbf{0.3441} & \textbf{0.3820} & 0.5880 & \textbf{0.5423} & \textbf{0.5642} & \textbf{0.6108} & 0.6032 \\
        & 4 & \textbf{0.4856} & \textbf{0.2807} & \textbf{0.3197} & 0.5882 & \textbf{0.5398} & \textbf{0.5629} & \textbf{0.6087} & 0.6032 \\
        \hline
        \end{tabular}
\caption{Evaluation metrics for each level in the classification hierarchy. Level 1 corresponds to Y02G; level 2 to Y02G10/00 and Y02G20/00; level 3 to Y02G10/10, Y02G10/20, Y02G20/10 and Y02G20/20; and level 4 to Y02G10/22 and Y02G10/24.}
\label{table:results_per_level}
\end{table}

\textbf{Performance for each class.} Table \ref{table:results_per_class} shows the results per class obtained for SBNN and SBHNN. In line with the results obtained in Table \ref{table:results_per_level}, classes that are on top of the classification hierarchy have better scores and as we get deeper in the hierarchy, both models struggle to maintain the performance. As mentioned in relation with the results of Table \ref{table:results_per_level}, this is to be expected due to the heavy class imbalance and to the class importance $\beta$. Another potential explanation is that the model currently only considers the title and the abstract. It is possible that, based solely on this information, it is challenging, if not impossible, to classify the patent so deep in the hierarchy. We believe that more input information could help with this issue. In the future work, we discuss an approach that could be used to overcome this limitation, i.e. enrich the input by some content from the description of patents. 

\begin{table}[htb!]
\centering
 \renewcommand{\arraystretch}{1.5}
 \begin{tabular}{|l|c|c c c|c c c|c|c|}
        \hline
        \multicolumn{1}{|c}{}  &
         \multicolumn{1}{|c|}{} &
        \multicolumn{3}{c|}{\textbf{macro-avg.}}  &
        \multicolumn{3}{c|}{\textbf{micro-avg.}}&
        \multicolumn{1}{c|}{\textbf{AUPRC}} & 
        \multicolumn{1}{c|}{\textbf{Accuracy}} \\
        Model & Class & hP & hR & hF1 & hP & hR & hF1 & &\\
        \hline
        \multirow{ 9}{*}{SBNN} & Y02G & \textbf{0.7173} & 0.7160 & \textbf{0.7166} & \textbf{0.7412} & \textbf{0.7412} & \textbf{0.7412} &  \textbf{0.7076} & \textbf{0.7412}\\
        & Y02G10/00 & \textbf{0.6145} & 0.5946 & 0.6043 & \textbf{0.6186} & 0.6000 & 0.6091 & 0.6710 & \textbf{0.6833} \\
        & Y02G10/10 & \textbf{0.6020} & 0.5661 & \textbf{0.5832} & \textbf{0.6078} & 0.5747 & 0.5908 & 0.6443 & \textbf{0.6596} \\
        & Y02G10/20 & 0.4097 & 0.3964 & 0.4029 & \textbf{0.6186} & 0.5936 & 0.6058 & 0.6665 & \textbf{0.6780} \\
        & Y02G10/22 & 0.3073 & 0.2973 & 0.3022 & \textbf{0.6186} & 0.5883 & 0.6031  & 0.6627 & \textbf{0.6780}\\
        & Y02G10/24 & 0.3073 & 0.2973 & 0.3022 & \textbf{0.6186} & 0.5932 & 0.6056  & 0.6658 & \textbf{0.6780}\\
        & Y02G20/00 & \textbf{0.6454} & 0.3853 & 0.4330 & \textbf{0.6382} & 0.5194 & 0.5727  & 0.6414 & \textbf{0.6913}\\
        & Y02G20/10 & \textbf{0.6865} & 0.3305 & 0.4030 & \textbf{0.6422} & 0.4949 & 0.5590 & 0.6229 & \textbf{0.6872} \\
        & Y02G20/20 & \textbf{0.4303} & 0.2569 & 0.2886 & \textbf{0.6382} & 0.5193 & 0.5726  & 0.6411 & \textbf{0.6913}\\
        \hline
        \multirow{ 9}{*}{SBHNN} & Y02G & 0.7096 & \textbf{0.7208} & 0.7131 & 0.7294 & 0.7294 & 0.7294 & 0.7057 & 0.7294 \\
        & Y02G10/00 & 0.5958 & \textbf{0.6184} & \textbf{0.6053} & 0.5979 & \textbf{0.6298} & \textbf{0.6134} & \textbf{0.6710} & 0.6584 \\
        & Y02G10/10 & 0.5904 & \textbf{0.5789} & 0.5824 & 0.5935 & \textbf{0.5936} & \textbf{0.5936} & \textbf{0.6447} & 0.6365 \\
        & Y02G10/20 & \textbf{0.6472} & \textbf{0.4451} & \textbf{0.4615} & 0.5981 & \textbf{0.6241} & \textbf{0.6108} & \textbf{0.6676} & 0.6537 \\
        & Y02G10/22 & \textbf{0.6729} & \textbf{0.3632} & \textbf{0.3970} & 0.5983 & \textbf{0.6196} & \textbf{0.6088} & \textbf{0.6645} & 0.6537 \\
        & Y02G10/24 & \textbf{0.4854} & \textbf{0.3338} & \textbf{0.3461} & 0.5981 & \textbf{0.6237} & \textbf{0.6106} & \textbf{0.6672} & 0.6537 \\
        & Y02G20/00 & 0.5381 & \textbf{0.4581} & \textbf{0.4747} & 0.5897 & \textbf{0.5862} & \textbf{0.5880}  & \textbf{0.6429} & 0.6731\\
        & Y02G20/10 & 0.5677 & \textbf{0.4215} & \textbf{0.4658} & 0.5915 & \textbf{0.5667} & \textbf{0.5788} & \textbf{0.6288} & 0.6679 \\
        & Y02G20/20 & 0.3587 & \textbf{0.3054} & \textbf{0.3165} & 0.5897 & \textbf{0.5861} & \textbf{0.5879} & \textbf{0.6428} & 0.6731\\
        \hline
        \end{tabular}
\caption{Evaluation metrics for each class in the classification hierarchy.}
\label{table:results_per_class}
\end{table}

\begin{figure}[htb!]
%\hspace*{-1in}
\centering
\noindent\includegraphics[width=1.1\columnwidth]
{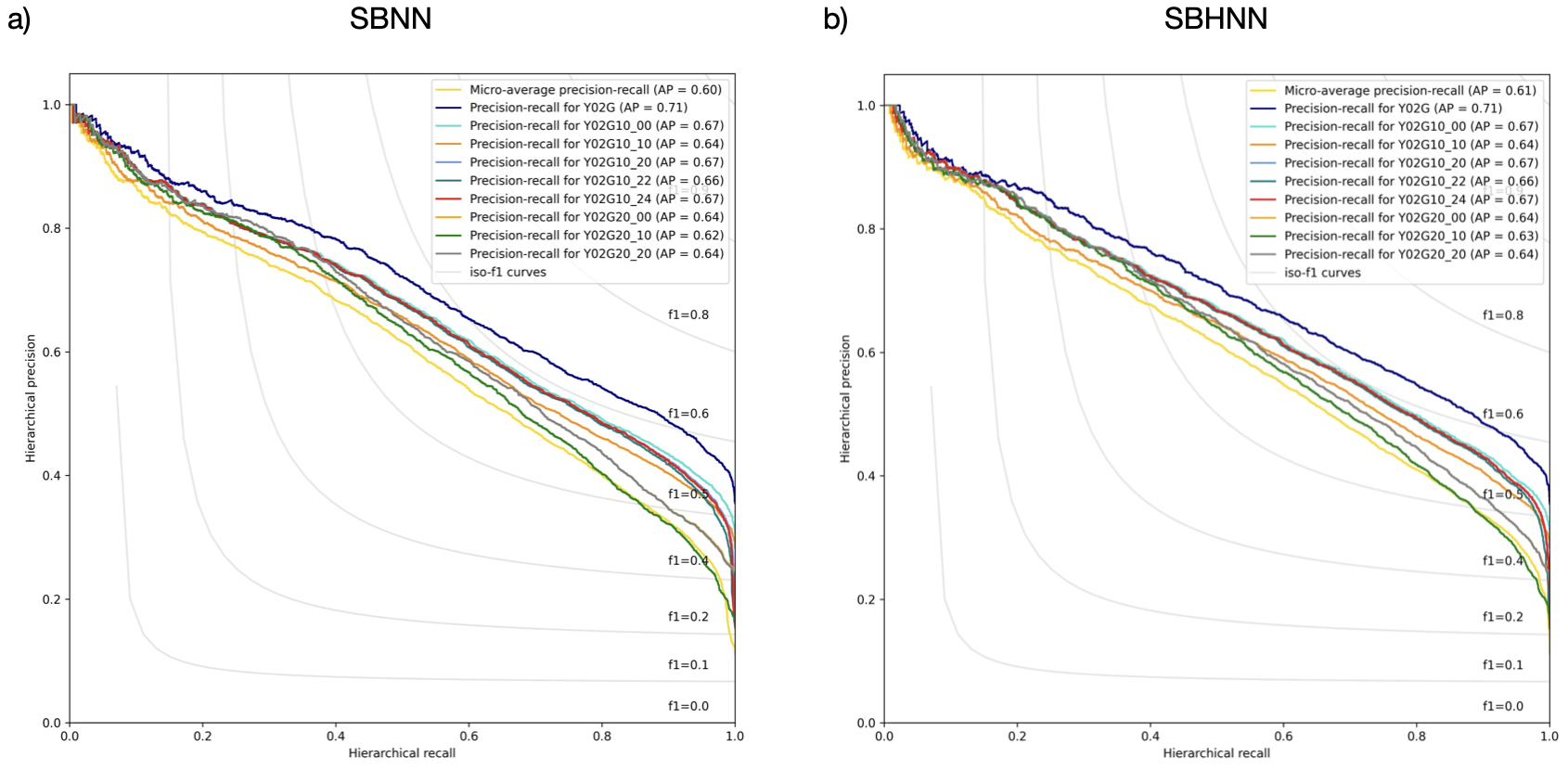}
\captionsetup{font={footnotesize}}
\protect\caption{\label{fig:pr_curve} Hierarchical precision - hierarchical recall curves for each class. Best viewed in color.}
\vspace{0.6cm}
\end{figure}

Figure \ref{fig:pr_curve} shows the hierarchical precision - hierarchical recall curves for each class when modifying the decision threshold (which for the results shown in Tables \ref{table:results_whole}, \ref{table:results_per_level} and \ref{table:results_per_class} was set to 0.5). These curves show how it is possible to trade off precision and recall by modifying the decision threshold. 
%This plots show us again the models' performance is decreasing when the class level goes deeper. 
%{EPO_codefest/fig/precision-recall-curve}

\subsection{Explainability}
\label{subsec:explainability}

\begin{figure}[htb!]
%\hspace*{-1in}
\centering
\noindent\includegraphics[width=1\columnwidth]{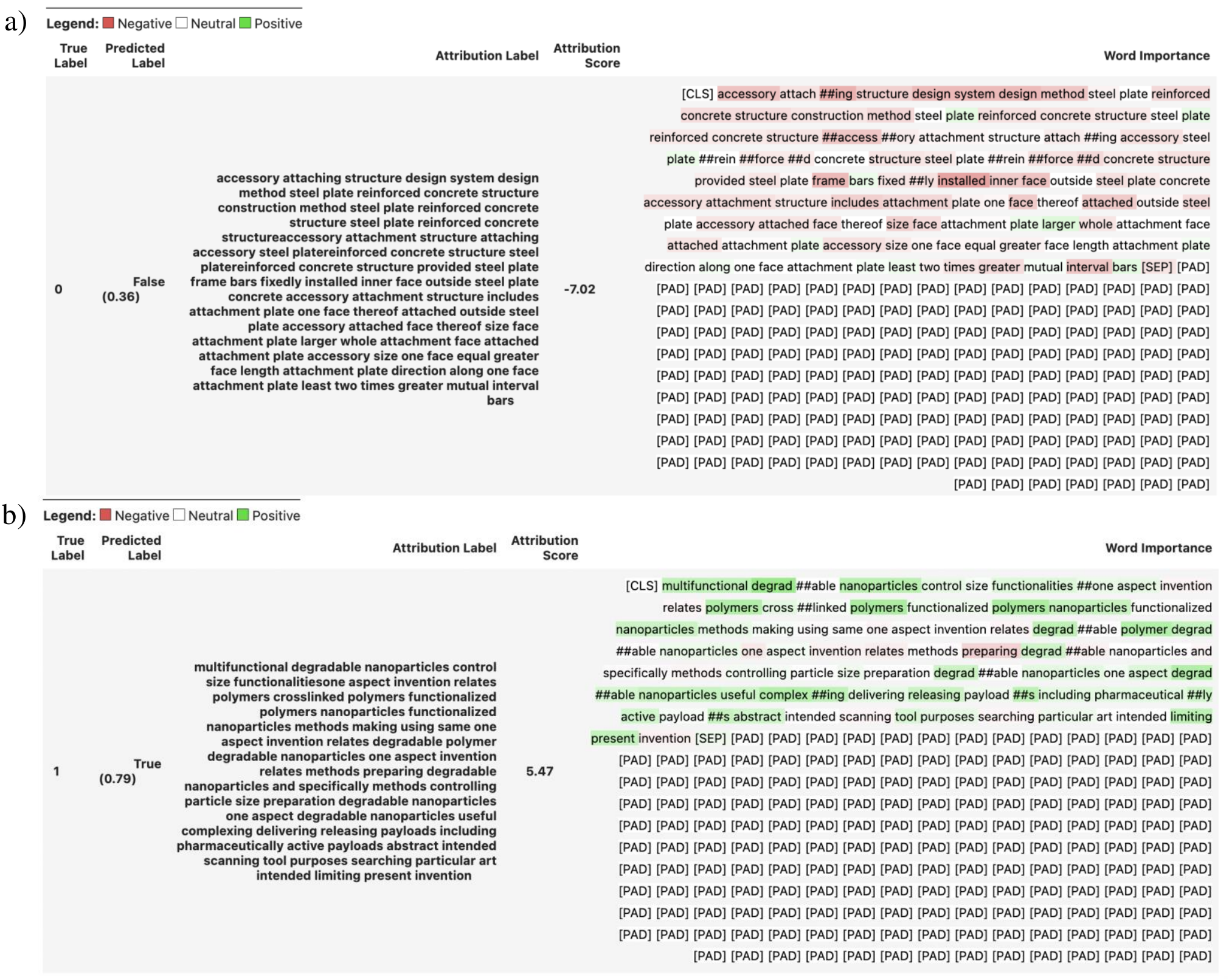}
\captionsetup{font={footnotesize}}
\protect\caption{\label{fig:visualization} a) Negative sample. b) Positive sample. Red means that the highlighted words decrease the probability of the patent being classified as belonging to green plastics whereas green means that the highlighted words increase the probability of the patent being classified as belonging to green plastics. We can also see how the BERT tokenizer tokenizes the input text. [CLS], [SEP] and [PAD] are special tokens of the BERT tokenizer. "$\#\#$" means that the rest of the token should be attached to the previous one, without space. These results were obtained with SBHNN. Best viewed in color.}
%\vspace{-0.45cm}
\end{figure}

There is an increasing interest in being able to explain the predictions of machine learning models. For this purpose, we implemented the method of integrated gradients \cite{DBLP:journals/corr/SundararajanTY17}, which allows us to find out which of the input words are the most important for the classification model when making a prediction. We implemented this method with the Captum library \cite{kokhlikyan2020captum}. In Figure \ref{fig:visualization}a, we show an example of a patent that does not relate to green plastics and in Figure \ref{fig:visualization}b, we show an example of a patent that relates to green plastics. 

This information is useful for understanding how the classification model is making the predictions. We believe that showing this information could help patent examiners gain trust in the classification model and also quickly decide whether the predictions of the classification model are reasonable or not.

\section{EPO CodeFest evaluation criteria}

In this section, we highlight how we believe that our solution fulfills the evaluation criteria.

\subsection{Completeness and transferability.} 

We have provided an end-to-end solution that covers all the steps, i.e. from pre-processing of the raw xml data to building a labeled training dataset, to training and evaluating classification models based on state-of-the-art machine learning algorithms. Together with the code repository in GitHub, we also submit slides to help people quickly understand our approach and this paper to present all the relevant technical details.

Our solution and code base are far from being limited to this specific problem of classifying patents relating to green plastics but can easily be applied to different situations. Whenever there is an update of the classification scheme, it is possible to use our code to construct a labeled training dataset and to train and evaluate a classification model for the updated classification scheme. Our code is very flexible and most modifications only need to be done in a configuration file. Moreover, it is also possible to skip the labeling process and to use our code to directly train a classification model with a given labeled training dataset for an already existing classification scheme. 

\subsection{Effectiveness and efficiency.} 

The effectiveness of the approach has been thoroughly discussed in Section \ref{subsection:experimental_resuls} in relation to the results of Tables \ref{table:results_whole}, \ref{table:results_per_level} and \ref{table:results_per_class}. The models' performance shows that our solution sets up a strong benchmark for this challenging problem. Possible ways that we believe could lead to still better results are discussed in the future work section. 

This approach can be applied whenever new classes are added to the classification scheme. Currently, this is done by humans which is a very time-consuming task and also prone to errors. With our solution, this could be done in a more automatic fashion by (re-)training a classification model for the updated classification scheme. Moreover, during inference, it takes our models less than a second to assign classes to a patent.

\subsection{Design and usability.} 

We have designed an end-to-end flexible pipeline to tackle this challenging problem. Moreover, we allow feedback from domain experts to be seamlessly incorporated (for example, by enriching the keywords associated with each class).

Our code repository is very well-structured and commented which makes it very easy to understand and use. Moreover, we follow industry best coding practices, e.g.: 1) The model results are reproducible and we provided all the pre-trained models which are ready to use; 2) we use popular open-sourced python packages, see the `requirements.txt` file to see all the packages' dependencies and corresponding versions, e.g. Pandas \cite{reback2020pandas} for raw data pre-processing, PyTorch \cite{NEURIPS2019_9015} for deep learning framework, HuggingFace \cite{huggingface}) for pre-trained BERT models; 3) the code is highly modular, there are no hard-coded variables and all the important parameters related to the projects are clearly defined in a configuration file; 4) we provided Jupyter Notebooks to help readers or users test the code and "play" with the pre-trained models; 5) we have this paper to document all the details related to this project and the ReadMe.md file in the code repository to help quickly set up the right python environment to test the code; and 6) the project can be installed as a python package and all of our functions can be imported and reused in any other projects.

\subsection{Creativity and innovation.} 

All of the research in patent classification focuses on an existing classification scheme. However, in our problem, we would like to classify patents in a classification scheme that does not exist yet. This is a very challenging problem and to the best of our knowledge, there is no prior art disclosing any solution. We made several creative design choices in order to solve this problem. We proposed to assign labels based on keyword matching to obtain a labeled dataset in order to train a machine learning classification model. We also proposed to have hierarchical labels in order to provide more positive training samples for the classification model. The two proposed models have novel structures and loss functions that are able to be trained for the complex multi-label hierarchical classification problem. 

We believe that our solution of assigning labels based on keyword matching to obtain a labeled dataset in order to learn a machine learning classification model is an innovative solution to the problem of re-classification of patents upon an update of the classification scheme. This solution can allow to speed up this laborious task. It could also be possible to use our solution in combination with the current human work. For example, our classification model could classify the patents and also provide an explanation as shown in Section \ref{subsec:explainability}. A human, e.g. a patent examiner, could quickly review the classes proposed by the classification model together with the explanation, and in many cases, probably directly take the proposed classes without the need to further review the patent.

\section{Conclusion}

We have provided an approach for the classification of patents related to green plastics based on an automatically obtained labeled training dataset. We have come up with two innovative and effective classification models and reported several evaluation metrics that give a complete overview of the performance of the models. Our models set a strong benchmark for this challenging and new problem. Our solution has great potential to improve the productivity of patent classification. Moreover, we highlight that this approach is not limited to the classification of green plastics, but it can be easily adapted for other fields whenever the classification hierarchy is updated. 

\section{Future work}

Here, we discuss some future work which, we believe, could help to further improve our solution. 

The next step would be to discuss the proposed classification scheme and the list of keywords assigned to each class with domain experts. Both the classification scheme and the list of keywords can be updated based on feedback. For example, as pointed out earlier, for classes such as Y02G10/24 or Y02G20/20, there are very few patents. It could be possible to either remove these classes from the classification scheme or to provide more relevant keywords to identify more patents belonging to those classes. Our code is very flexible and both modifications can be implemented almost effortlessly.

An interesting experiment to double-check the performance of our models could be to manually build a test dataset with samples from all classes in the classification hierarchy and to evaluate the performance of the model.

In the future, as incoming patents will be manually labeled in these classes, it would also be possible to periodically update the model with the newly labeled data. It would then be possible to give higher importance to the manually labeled patents in comparison to patents having automatically assigned labels. 

So far, we have limited our approach to dealing with patents written in English. In order to be able to also classify patents in French and German, it is possible to re-use this code to build a specific classification model for each language. Alternatively, we saw that, recently, a pre-trained language-agnostic BERT model has been open-sourced \cite{feng-etal-2022-language}, so it could be interesting to explore the possibility of using such a model to process all three languages. 

Most of the prior-art works \cite{li2018deeppatent, huang2019hierarchical, lee2020patent, pujari2021multi} use the title and the abstract as input for the classification. However, patent examiners usually classify patents based on the content of the description. Since the descriptions are usually at least a couple of pages long, it is not straightforward to obtain a proper embedding using a transformer model such as BERT. It could be possible to try some of the approaches proposed in \cite{park2022efficient} so that the content of the description is also taken into account.

{\small
\bibliographystyle{ieee}
\bibliography{neurips_2019}

\begin{thebibliography}{10}\itemsep=-1pt

\bibitem{beltagy-etal-2019-scibert}
I.~Beltagy, K.~Lo, and A.~Cohan.
\newblock {S}ci{BERT}: A pretrained language model for scientific text.
\newblock In {\em Proceedings of the 2019 Conference on Empirical Methods in
  Natural Language Processing and the 9th International Joint Conference on
  Natural Language Processing (EMNLP-IJCNLP)}, pages 3615--3620, Hong Kong,
  China, Nov. 2019. Association for Computational Linguistics.

\bibitem{weak_super}
L.-M. Chen, B.-X. Xiu, and Z.-Y. Ding.
\newblock Multiple weak supervision for short text classification.
\newblock {\em Applied Intelligence}, 52(8):9101–9116, jun 2022.

\bibitem{davis2006relationship}
J.~Davis and M.~Goadrich.
\newblock The relationship between precision-recall and roc curves.
\newblock In {\em Proceedings of the 23rd international conference on Machine
  learning}, pages 233--240, 2006.

\bibitem{devlin2018bert}
J.~Devlin, M.-W. Chang, K.~Lee, and K.~Toutanova.
\newblock Bert: Pre-training of deep bidirectional transformers for language
  understanding.
\newblock {\em arXiv preprint arXiv:1810.04805}, 2018.

\bibitem{d2013text}
E.~D'hondt, S.~Verberne, C.~Koster, and L.~Boves.
\newblock Text representations for patent classification.
\newblock {\em Computational Linguistics}, 39(3):755--775, 2013.

\bibitem{fall2003automated}
C.~J. Fall, A.~T{\"o}rcsv{\'a}ri, K.~Benzineb, and G.~Karetka.
\newblock Automated categorization in the international patent classification.
\newblock In {\em Acm Sigir Forum}, volume~37, pages 10--25. ACM New York, NY,
  USA, 2003.

\bibitem{feng-etal-2022-language}
F.~Feng, Y.~Yang, D.~Cer, N.~Arivazhagan, and W.~Wang.
\newblock Language-agnostic {BERT} sentence embedding.
\newblock In {\em Proceedings of the 60th Annual Meeting of the Association for
  Computational Linguistics (Volume 1: Long Papers)}, pages 878--891, Dublin,
  Ireland, May 2022. Association for Computational Linguistics.

\bibitem{guyot2010myclass}
J.~Guyot, K.~Benzineb, G.~Falquet, and S.~Shift.
\newblock myclass: A mature tool for patent classification.
\newblock In {\em CLEF (Notebook papers/LABs/workshops)}, 2010.

\bibitem{huang2019hierarchical}
W.~Huang, E.~Chen, Q.~Liu, Y.~Chen, Z.~Huang, Y.~Liu, Z.~Zhao, D.~Zhang, and
  S.~Wang.
\newblock Hierarchical multi-label text classification: An attention-based
  recurrent network approach.
\newblock In {\em Proceedings of the 28th ACM international conference on
  information and knowledge management}, pages 1051--1060, 2019.

\bibitem{kim2008visualization}
Y.~G. Kim, J.~H. Suh, and S.~C. Park.
\newblock Visualization of patent analysis for emerging technology.
\newblock {\em Expert systems with applications}, 34(3):1804--1812, 2008.

\bibitem{kingma2014adam}
D.~P. Kingma and J.~Ba.
\newblock Adam: A method for stochastic optimization.
\newblock {\em arXiv preprint arXiv:1412.6980}, 2014.

\bibitem{kiritchenko2005functional}
S.~Kiritchenko, S.~Matwin, A.~F. Famili, et~al.
\newblock Functional annotation of genes using hierarchical text
  categorization.
\newblock In {\em Proc. of the ACL Workshop on Linking Biological Literature,
  Ontologies and Databases: Mining Biological Semantics}, 2005.

\bibitem{kokhlikyan2020captum}
N.~Kokhlikyan, V.~Miglani, M.~Martin, E.~Wang, B.~Alsallakh, J.~Reynolds,
  A.~Melnikov, N.~Kliushkina, C.~Araya, S.~Yan, and O.~Reblitz-Richardson.
\newblock Captum: A unified and generic model interpretability library for
  pytorch, 2020.

\bibitem{lee2020patent}
J.-S. Lee and J.~Hsiang.
\newblock Patent classification by fine-tuning bert language model.
\newblock {\em World Patent Information}, 61:101965, 2020.

\bibitem{li2018deeppatent}
S.~Li, J.~Hu, Y.~Cui, and J.~Hu.
\newblock Deeppatent: patent classification with convolutional neural networks
  and word embedding.
\newblock {\em Scientometrics}, 117(2):721--744, 2018.

\bibitem{office}
E.~P. Office.
\newblock Ep full-text data for text analytics.

\bibitem{reback2020pandas}
T.~pandas~development team.
\newblock pandas-dev/pandas: Pandas, Feb. 2020.

\bibitem{park2022efficient}
H.~H. Park, Y.~Vyas, and K.~Shah.
\newblock Efficient classification of long documents using transformers.
\newblock {\em arXiv preprint arXiv:2203.11258}, 2022.

\bibitem{NEURIPS2019_9015}
A.~Paszke, S.~Gross, F.~Massa, A.~Lerer, J.~Bradbury, G.~Chanan, T.~Killeen,
  Z.~Lin, N.~Gimelshein, L.~Antiga, A.~Desmaison, A.~Kopf, E.~Yang, Z.~DeVito,
  M.~Raison, A.~Tejani, S.~Chilamkurthy, B.~Steiner, L.~Fang, J.~Bai, and
  S.~Chintala.
\newblock Pytorch: An imperative style, high-performance deep learning library.
\newblock In {\em Advances in Neural Information Processing Systems 32}, pages
  8024--8035. Curran Associates, Inc., 2019.

\bibitem{scikit-learn}
F.~Pedregosa, G.~Varoquaux, A.~Gramfort, V.~Michel, B.~Thirion, O.~Grisel,
  M.~Blondel, P.~Prettenhofer, R.~Weiss, V.~Dubourg, J.~Vanderplas, A.~Passos,
  D.~Cournapeau, M.~Brucher, M.~Perrot, and E.~Duchesnay.
\newblock Scikit-learn: Machine learning in {P}ython.
\newblock {\em Journal of Machine Learning Research}, 12:2825--2830, 2011.

\bibitem{EPOgreenplastics}
J.~Pose-Rodriguez, Y.~Ménière, I.~Rudyk, M.~Dossin, M.~Grilli, D.~Marsitzky,
  W.~Meiser, J.~Philpott, C.~Rossatto, F.~Tassinari, P.~Vandoolaeghe, and
  S.~Wewege.
\newblock Patents for tomorrow's plastics global innovation trends in
  recycling, circular design and alternative sources.
\newblock 10 2021.

\bibitem{pujari2021multi}
S.~C. Pujari, A.~Friedrich, and J.~Str{\"o}tgen.
\newblock A multi-task approach to neural multi-label hierarchical patent
  classification using transformers.
\newblock In {\em European Conference on Information Retrieval}, pages
  513--528. Springer, 2021.

\bibitem{DBLP:journals/corr/SundararajanTY17}
M.~Sundararajan, A.~Taly, and Q.~Yan.
\newblock Axiomatic attribution for deep networks.
\newblock {\em CoRR}, abs/1703.01365, 2017.

\bibitem{trappey2006development}
A.~J. Trappey, F.-C. Hsu, C.~V. Trappey, and C.-I. Lin.
\newblock Development of a patent document classification and search platform
  using a back-propagation network.
\newblock {\em Expert Systems with Applications}, 31(4):755--765, 2006.

\bibitem{huggingface}
T.~Wolf, L.~Debut, V.~Sanh, J.~Chaumond, C.~Delangue, A.~Moi, P.~Cistac,
  T.~Rault, R.~Louf, M.~Funtowicz, J.~Davison, S.~Shleifer, P.~von Platen,
  C.~Ma, Y.~Jernite, J.~Plu, C.~Xu, T.~L. Scao, S.~Gugger, M.~Drame, Q.~Lhoest,
  and A.~M. Rush.
\newblock Huggingface's transformers: State-of-the-art natural language
  processing, 2019.

\bibitem{wu2010patent}
C.-H. Wu, Y.~Ken, and T.~Huang.
\newblock Patent classification system using a new hybrid genetic algorithm
  support vector machine.
\newblock {\em Applied Soft Computing}, 10(4):1164--1177, 2010.

\end{thebibliography}
}
\end{document}